\documentclass[a4paper,USenglish,cleveref,autoref,thm-restate]{lipics-v2021}

\usepackage[super]{nth}

\title{Learning a Generic Value-Selection Heuristic Inside a Constraint Programming Solver} 

\titlerunning{Learning a Generic Value-Selection Heuristic}

\author{Tom {Marty}}{Polytechnique Montréal, Montreal, Canada \and Ecole Polytechnique, Palaiseau, France \and \url{https://3rdcore.github.io/} }{tom.marty@polymtl.ca}{https://orcid.org/0009-0001-3468-3327}{}

\author{Tristan {François}}{Ecole Polytechnique, Palaiseau, France}{tristan.francois@polytechnique.edu}{}{}

\author{Pierre {Tessier}}{Ecole Polytechnique, Palaiseau, France}{pierre.tessier@polytechnique.edu}{}{}

\author{Louis {Gautier}}{Ecole Polytechnique, Palaiseau, France}{Louis.Gautier@polytechnique.edu}{}{}

\author{Louis-Martin {Rousseau}}{Polytechnique Montréal, Montreal, Canada \and \url{https://www.polymtl.ca/expertises/rousseau-louis-martin} }{louis-martin.rousseau@polymtl.ca}{https://orcid.org/0000-0001-6949-6014}{}

\author{Quentin {Cappart}}{Polytechnique Montréal, Montreal, Canada \and \url{https://qcappart.github.io/}}{quentin.cappart@polymtl.ca}{https://orcid.org/0000-0002-8742-0774}{}

\authorrunning{T. Marty et al.} 

\Copyright{Marty, François, Tessier, Gautier, Rousseau, Cappart} 



\category{} 

\relatedversion{} 

\funding{The research presented in this paper has been partially funded thanks to a NSERC Discovery Grant held by Quentin Cappart.}

\acknowledgements{We thank CIRRELT for providing essential computing resources. We extend our gratitude to all the people who contributed to the related open-source project: Léo Boisvert, Tom Sander, Ziad El Assal, Malik Attalah, and Marco Novaes.}

\nolinenumbers 

\EventEditors{Roland H. C. Yap}
\EventNoEds{1}
\EventLongTitle{29th International Conference on Principles and Practice of Constraint Programming (CP 2023)}
\EventShortTitle{CP 2023}
\EventAcronym{CP}
\EventYear{2023}
\EventDate{August 27--31, 2023}
\EventLocation{Toronto, Canada}
\EventLogo{}
\SeriesVolume{280}
\ArticleNo{4}

\begin{document}

\maketitle

\begin{abstract}
Constraint programming  is known for being an efficient approach to solving combinatorial problems. Important design choices in a solver are the \textit{branching heuristics}, designed to lead the search to the best solutions in a minimum amount of time. However, developing these heuristics is a time-consuming process that requires problem-specific expertise. This observation has motivated many efforts to use machine learning to automatically learn efficient heuristics without expert intervention. Although several generic \textit{variable-selection heuristics} are available in the literature,  the options for \textit{value-selection heuristics} are more scarce.  We propose to tackle this issue by introducing a generic learning procedure that can be used to obtain a value-selection heuristic inside a constraint programming solver. This has been achieved thanks to the combination of a \textit{deep Q-learning} algorithm, a tailored \textit{reward signal}, and a \textit{heterogeneous graph neural network}. Experiments on \textit{graph coloring}, \textit{maximum independent set}, and \textit{maximum cut} problems show that this framework 
competes with the well-known impact-based and activity-based search heuristics and can find solutions close to optimality without requiring a large number of backtracks.
\end{abstract}

\section{Introduction}
\label{sec:intro}

Combinatorial optimization has countless industrial applications, such as scheduling, routing, or finance. Unfortunately, most of these problems are NP-hard and, thereby, challenging to solve efficiently. It is why finding good solutions has motivated intense research efforts for many years. Traditional methods for tackling them are somehow based on a \textit{search procedure}: A clever enumeration of the solution space is performed to find a feasible and possibly optimal solution. Among these methods, \textit{constraint programming} (CP) is an exact procedure. It constitutes a popular approach as it offers the possibility to find the optimal solution or good feasible approximations by stopping the search early. An additional asset is its declarative paradigm in modeling, which makes the technology easier for the end-user to grasp. Introducing solver-agnostic modeling languages, such as MiniZinc~\cite{nethercote2007minizinc} has greatly facilitated this aspect. Aligned with this goal, the propagation engine inside a CP solver is mostly hidden from the end-user. However, ensuring a generic search procedure is trickier as non-trivial heuristics must be designed to make the solving process efficient for an arbitrary problem. That being said, generic \textit{variable-selection} and \textit{value-selection} heuristics have been successfully designed. Notable examples are \textit{impact-based search}~\cite{refalo2004impact} or \textit{activity-based search}~\cite{michel2012activity}, but they require computationally intensive initialization and yield poor performance at the beginning of the search. This makes these methods not always appropriate for general use. As a concrete example, the current version of MiniZinc\footnote{https://www.minizinc.org/doc-2.7.0/en} does not propose generic value-selection heuristics, except \textit{in(out)domain} or \textit{impact-based search}. In practice, heuristics are often designed thanks to problem-specific expert knowledge, which is often out of reach for end-users that do not have a solid background in artificial intelligence.

In another context, \textit{machine learning} (ML) has been recently considered for automating the design of branching heuristics, both in constraint programming \cite{cappart_combining_2020}, mixed-integer programming \cite{gasse_exact_2019,khalil_learning_2016}, or SAT solving~\cite{selsam2019guiding}.
Specifically, \textit{reinforcement learning}  (RL)~\cite{sutton2018reinforcement} or \textit{imitation learning}~\cite{hussein2017imitation} approaches, often combined with \textit{deep learning}~\cite{lecun2015deep}, have gained special attention.
Although this idea seems appealing, this is not an easy task to achieve in practice as several technical considerations must be taken into account in
order to ensure both the efficiency and the genericity of the approach. 
In constraint programming, we identified three questions to resolve when learning a generic branching heuristic inside a solver. They are as follows:
\begin{enumerate}
    \item \textit{How to train the machine learning model?} An intuitive way is to leverage an RL agent that would explore the tree search by making branching decisions and rewarding it based on the quality of the solution found on a terminal node. 
    This would typically be done with a \textit{depth-first search} traversal of the tree for getting a certificate of optimality. 
    However, as pointed out by several authors \cite{treeMDP,song_learning_2022}, the backtracking operations inside a solver raise difficulties when formalizing the task as a Markov decision process and may require redefining it. 
    Besides, this training scheme intensifies the \textit{credit assignment problem} \cite{4066245}, ubiquitous in reinforcement learning. 
    \item \textit{How to evaluate the quality of a value selection?} A core component of an RL environment is the \textit{reward function}, which gives a score to each decision performed. The end goal for the agent is to perform a sequence of decisions leading to the best-accumulated sum of rewards. In our case, an intuitive solution would be to reward the agent according to the quality of the solution found. 
    However, this information is only available at terminal nodes, and only a zero reward is provided in branching nodes. This is related to the \textit{sparse reward} problematic, which is known to complicate the training process.
    \item \textit{How to learn from a CP model?} This question relates to the type of architecture that can obtain a value-selection heuristic from a search node (i.e., a partially solved CP model). A promising direction has been proposed by Gasse et al. \cite{gasse_exact_2019} for binary mixed-integer programs. They introduced a bipartite graph linking variables and constraints (i.e., the two types of nodes) when a variable is involved in a given constraint. The subsequent architecture is a \textit{heterogeneous graph neural network}. However, this encoding is not directly applicable in constraint programming, as a CP model generally involves non-binary variables and combinatorial constraints.
     This has been partially addressed by Chalumeau et al.~\cite{chalumeau_seapearl_2021}, who introduced a tripartite graph where variables, values, and constraints are specific types of nodes. However, this approach lacks genericity as the method requires retraining when the number of variables changes. 
\end{enumerate}
To our knowledge, answering such questions is still an open challenge in the research community. This paper proposes to progress in this direction. It introduces a generic learning procedure that can be used to obtain a value-selection heuristic from a constraint programming model given as input.
The approach has been designed to be generic in that it can be used for any CP model given as input. In practice, a specific way to extract \textit{features} from a constraint should be designed for any available constraint, but this has to be done only once per constraint type. In this proof of concept, we limit our experiments to three combinatorial optimization problems, namely \textit{graph coloring}, \textit{maximum independent set}, and \textit{maximum cut}.
Specifically, we propose three main contributions, each dedicated to addressing one of the aforementioned difficulties.
They are as follows: (1) a learning procedure, based on restarts, for training a reinforcement learning agent directly inside a CP solver,
(2) a reward function able to assign non-zero intermediate rewards based on the propagation that has been carried out on the node, 
and (3) a neural architecture based on a tripartite graph representation and a heterogeneous graph neural network.
Experimental results show that combining these three ideas enables the search to find good solutions without requiring many backtracks and competes with the well-known impact-based and activity-based search heuristics. 

The paper is structured as follows. The next section presents other approaches related to our contribution.
Then, Section \ref{sec:TB} introduces succinctly technical background on reinforcement learning and graph neural networks.
The core contributions are then presented in Section \ref{sec:contrib}.
Finally, Section \ref{sec:exp} provides experimental results and closes with a discussion of the results.

\section{Related Work}
\label{sec:rel_work}

Bengio et al.~\cite{bengio_machine_2020} identified three ways to leverage machine learning for combinatorial optimization.
First, \textit{end-to-end} learning aims to solve the problem only with a trained ML model. This has been, for instance, considered for
the traveling salesman problem \cite{bello_neural_2016,kool2018attention}. However, such an approach does not guarantee the validity nor optimality of the solution obtained. Second, \textit{learning to configure} is dedicated
to providing insights to a solver before its execution. This can be, for instance, the decision to linearize the problem in the context of quadratic programs \cite{bonami2018learning} or to learn when a decomposition is appropriate \cite{kruber2017learning}.
This approach is also referred to as parameter tuning~\cite{hoos2011automated}. 
We refer to the initial survey for extended information about these two families of approaches.
Third, \textit{learning within a search procedure} uses machine learning within the solver.
Our contribution belongs to this last category of methods.
Although the idea of combining learning and searching for solving combinatorial optimization problems was already discussed in the nineties~\cite{potvin1996hybrid}, it has re-emerged recently with the rise of deep learning. 
Most combinatorial optimization solvers are based on \textit{branch-and-bound} and \textit{backtracking}.
In this context, ML is often used with branching rules to follow. \textit{Imitation learning}~\cite{hussein2017imitation} 
has been for instance used to replicate the expensive \textit{strong branching} strategy for mixed-integer programming solvers \cite{gasse_exact_2019,khalil_learning_2016}.
One limitation of imitation learning is that the performances are bounded by the performance of the imitated strategy, which remains heuristic
and perfectible~\cite{sun2020improving}.
This opens the door for RL approaches that have the guarantee to find the best branching strategy eventually~\cite{mazyavkina2021reinforcement}.
A branching strategy can be split into two challenging decisions, \textit{variable-selection} and \textit{value-selection}.
Reinforcement learning approaches have been considered for both of them.

Concerning the learning for selecting the next variable to branch on, Song et al. \cite{song_learning_2022} proposed to combine
a \textit{double deep Q-network} algorithm \cite{van2016deep} with a graph neural network for carrying out this task.
The approach is trained to minimize the expected number of nodes to reach a leaf node using the \textit{first-fail principle}. 
Although this is a good proxy for pruning a maximum of infeasible solutions for a constraint satisfaction problem,
it does not extend naturally to optimization variants, for which one should consider a trade-off between the quality of the solution found and the number of nodes required to reach that solution. 
Similarly, van Driel et al.~\cite{van2021learning} leveraged a graph neural network to initialize a variable-selection heuristic
for \textit{Chuffed}, a hybrid CP-SAT solver. In an online setting,
Doolard and Yorke-Smith~\cite{doolaard2022online} also proposed to learn variable ordering heuristics where training time is included in the total solving time.
Bandit-based learning approaches were also considered  by
Xia and Yap  to automatically select search heuristics~\cite{xia2018learning}. 

For the value-selection heuristic, 
Chu and Stuckey~\cite{chu2015learning} introduced a scoring function which gives a score indicating how good an assignation is,
given the current domain. A training phase is the carried out in a supervised manner
to learn this scoring function. Cappart et al.~\cite{cappart_combining_2020} proposed to train a model with reinforcement learning outside the CP solver and to integrate the agent, once trained, subsequently in the solver. This has been achieved by reaping the benefits of a dynamic programming formulation of a combinatorial problem.
An important limitation of this work is that no information related to the CP solver, such as the propagation achieved on a node, can be used to drive the decision. Chalumeau et al.~\cite{chalumeau_seapearl_2021} mitigated this issue by carrying out the learning inside the solver.
The model is trained to find the optimal solution and to prove it with the least number of explored search nodes.
However, this goal is disconnected from finding the best solution as quickly as possible and is practically hard to achieve, even with a good heuristic.
A more realistic goal is to find a good solution quickly without closing the search.
This is how the contribution of this paper is positioned.

We want to point out that learning \textit{how to branch} is not the only way to leverage ML inside a combinatorial optimization solver. 
Related works have also been proposed on learning tight optimization bounds~\cite{cappart2022improving} or for accelerating
column generation approaches~\cite{morabit2021machine}. 
A recurrent design choice is an architecture based on graph neural networks. We refer to the following survey for
more information about combinatorial optimization with graph neural networks~\cite{cappart_GNN}.

\section{Technical Background}
\label{sec:TB}
This section introduces the required  background on reinforcement learning
and graph neural network to grasp the technical aspects of the paper.
 
\subsection{Reinforcement Learning}
\label{sec:RL}
Let  $\langle S,A,T,R \rangle$ be a 4-tuple representing a \textit{Markov decision process} where $S$ is the
set of states in the environment, $A$ is the set of actions that the agent can do, $T : S \times A \to S$
is a transition function leading the agent from one state to another, given the action taken, and
$R : S \times A \to \mathbb{R}$ is a reward function of taking an action from a specific state. 
The sequence $[s_1,\dots,s_T]$ from the initial state ($s_1$) of an agent towards a terminal state ($s_T$) is referred to as an \textit{episode}.
The returned reward within a partial episode $[s_t,\dots,s_T]$ can be formalized as follows:  $G_t = \sum_{i=t}^T R(s_i,a_i)$.
We intentionally omitted the discounting factor as we do not want to discount the late rewards in our application.
The agent is governed by a policy $\pi : S \to A$, which indicates the action that must be taken on a given state.
The agent's goal is to find the policy that will lead it to maximize the accumulated reward until a terminal state is reached.
The core idea of reinforcement learning is to determine this policy by letting the agent interact with the environment and increasing the probability of taking action if it leads to high subsequent rewards.
There are a plethora of reinforcement learning algorithms dedicated to this task, such as \textit{trust region policy optimization}~\cite{schulman2015trust} or \textit{soft actor-critic}~\cite{haarnoja2018soft}. We refer to \textit{SpinningUp} website for explanations of the main algorithms~\cite{SpinningUp2018}.

This section presents the core principles of \textit{deep Q-learning}~\cite{DQN}, which is the algorithm used in this paper.
The idea is to compute an \textit{action-value} function $Q^{\pi}(s_t, a_t) = G_t$. Intuitively, this function gives
the accumulated reward that the agent will obtain when performing the action $a$ at state $s$ while subsequently following a policy $\pi$.
The output of this function for a specific action is referred to as a \textit{Q-value}.
Provided that the action-value function can be computed exactly, the optimal policy $\pi^\star$ turns to be simply the selection of the action having the highest Q-value on a specific state: $\pi^* = \mathsf{argmax}_\pi Q^{\pi}(s,a), \ \forall (s,a) \in (S,A)$. 
Although the exact computation of $Q$-values can theoretically be performed, 
a specific value must be computed for each pair of states and actions, which is not tractable for realistic situations.
It is why a tremendous amount of work has been carried out to approximate accurately and efficiently $Q$-values.
Among them, deep $Q$-learning aims to provide a neural estimator $\hat{Q}(s,a,\theta) \approx Q(s,a)$, where $\theta$ is a tensor of parameters that
must be learned during a training phase. This algorithm is commonly enriched with other mechanisms dedicated to speed-up or stabilizing the training process,
such as the \textit{double deep Q-network} variant~\cite{van2016deep} or
\textit{prioritized experience replay}~\cite{schaul2015prioritized}.
Concerning the neural architecture, we opted for a graph neural network, which is explained in the next section.

\subsection{Graph Neural Network}
\label{sec:gnn}
Intuitively, the goal of a \textit{graph neural network} (GNN) is to embed information contained
in a graph (e.g., the structure of the graph, spatial properties, 
features of the nodes, etc.) into a $d$-dimensional tensor for each node $u \in V$ of the graph.
To do so, information on a node is iteratively refined by aggregating information from neighboring nodes.
Each iteration of aggregation is referred to as a \textit{layer} of the GNN and involves parameters that must be learned.
Let $h_u^{k} \in \mathbb{R}^{d\times 1}$ be the tensor representation of node $u$ at layer $k$ of the GNN,  
 $h_u^{k+1} \in \mathbb{R}^{l \times 1}$ be the tensor representation of this node at the next layer ($l$ being the dimension of a node at the layer $k+1$), and
 $\theta_1 \in  \mathbb{R}^{l \times d}$ and $\theta_2 \in \mathbb{R}^{l \times d}$ be two matrices of parameters, respectively.
Each GNN layer carries out the following update:
\begin{equation}
\label{eq:gnn}
    h_{u}^{k+1} = g\Big(\theta_1 h_u^{k}  ~ \star ~ (\bigoplus_{v \in N(u)} \theta_2 h_v^{k}) \Big)  ~ ~ \forall{u \in V}
\end{equation}
Three operations are involved in this update:
(1) $\bigoplus$ is an \textit{aggregation} operator that is dedicated to aggregating the information of neighbors (e.g., \textit{mean-pooling} or \textit{sum-pooling}),
(2) $\star$ is a \textit{merging} which enables to combine of the information of a node with the ones from the neighbors
(e.g., a concatenation),
and (3) $g$ is an element-wise non-linear activation function, such as the ones commonly used in fully-connected neural networks (e.g., \textsf{ReLU}~\cite{glorot2011deep}). Without loss of generality, the \textit{bias} term is not included in the equation.
A concrete implementation of a GNN defines these three functions adequately.
The training is conducted in a fully-connected neural network through  back-propagation and an optimizer based on stochastic gradient descent such as Adam~\cite{kingma2014adam}.

\section{Learning a Value-Selection Heuristic Inside a Solver}
\label{sec:contrib}

This section presents how a value-selection heuristic can be learned with reinforcement learning
in a CP solver from a  model given as input.
This is the core contribution of the paper. Three mechanisms are introduced:
(1) \textit{a training procedure based on restarts}, (2) \textit{a reward function leveraging propagation of domains}, and
(3) \textit{a heterogeneous graph neural network architecture}. 
They are described individually in the next subsections. They have been implemented in the recently introduced \textit{SeaPearl.jl} solver \cite{chalumeau_seapearl_2021}.
Inspired by the architecture of MiniCP~\cite{michel2021minicp},
the main specificity of SeaPearl is to natively integrate support for learning inside the search procedure. This greatly facilitates the prototyping of new search algorithms based on learning.

\subsection{Restart-Based Training}
\label{subsec:rbs}
Generally speaking, the performance of a reinforcement learning agent is tightly correlated with the definition of an \textit{episode}. This corresponds to the agent's interactions with the CP solver's search procedure and is related to the goal desired for the agent. Two options are discussed in this section, (1) an \textit{episode based on depth-first search},  introduced by Chalumeau et al.~\cite{chalumeau_seapearl_2021}, and (2) an \textit{episode based on restarts}, which is our first contribution.

\begin{figure}[!ht]
\centering
\includegraphics[width=\textwidth]{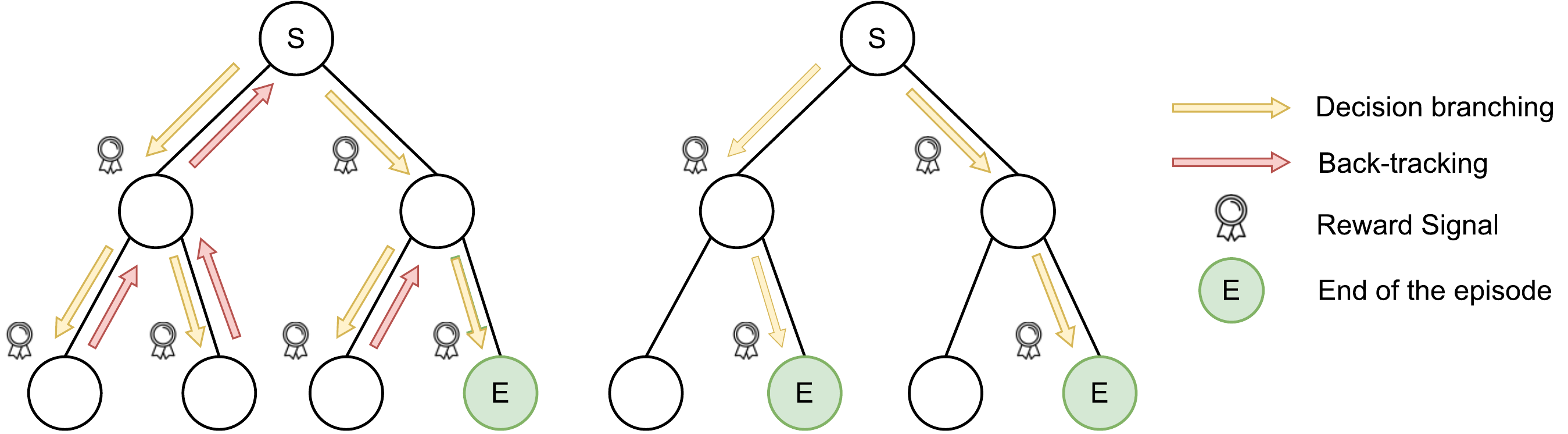}
\caption{The two training procedures (left: \textit{depth-first search}~\cite{chalumeau_seapearl_2021}, right: \textit{restart-based} - ours)}
\label{fig:rbs_dfs}
\end{figure}

Building branching heuristics for solving exact combinatorial optimization problems often concurrently targets two objectives: \textit{finding quickly good solutions} and \textit{proving the optimality of a solution}. The approach of Chalumeau et al.~\cite{chalumeau_seapearl_2021} relies heavily on the second objective and aims to minimize the number of visited search nodes before proving optimality (e.g., closing the search). To do so, they defined a training episode as a complete solving process carried out by  the depth-first search of a solver and penalized through the reward function the generation of each node.  This is illustrated in the left picture of Figure~\ref{fig:rbs_dfs}. However, this approach suffers from an important difficulty. An episode only terminates when the search is completed, which is often intractable for realistic problems as it requires exploring an exponentially large search tree. This is especially problematic during training, where the heuristic is still mediocre. In addition, using a depth-first search algorithm in a Markov Decision Process (MDP) framework required additional considerations not considered by Chalumeau et al.~\cite{chalumeau_seapearl_2021}. For example, using a backtracking algorithm in a regular temporal MDP renders their method prone to the \textit{credit assignment problem}~\cite{4066245}.  These considerations have been pointed out by Scavuzzo et al.~\cite{treeMDP} for mixed-integer programming.

Unlike this approach, we propose to train the model to find high-quality solutions quickly. To do so, we followed the approach proposed by Cappart et al.~\cite{cappart_combining_2020}: an episode is defined as a \textit{single dive} in the search tree. No backtrack is allowed; the episode stops when a complete solution is found or a failure is generated. Once the episode is terminated, a restart from the root node is performed, and a new episode is generated, hence the name of \textit{restart-based episode}.  This is illustrated in the right picture of Figure~\ref{fig:rbs_dfs}. One limitation of Cappart et al.~\cite{cappart_combining_2020} is that episodes are executed outside the CP solver during the training and cannot use the information updated during propagation for the branching. 
Inspired by Song et al.~\cite{song_learning_2022} for variable-selection heuristics, we addressed this limitation by executing each episode inside the solver during the training. Formally, this requires defining the dynamics of the environment as a Markov Decision Process (i.e., a tuple $\langle S,A,T,R \rangle$, see Section \ref{sec:RL}).
It is defined as follows.

\begin{description}
\item[Set of states] Let $\mathcal{P} = \langle X, D(X), C, O \rangle$ be the expression of a combinatorial optimization problem (COP), defined by its variables ($X$), the related domains  ($D$), its constraints ($C$), and an objective function ($O$). Each state $s_t \in S$ is defined as the pair $s_t = (\mathcal{P}_t,x_t)$, where $\mathcal{P}_t$ is a partially solved COP (i.e., some variables may have been assigned), and $x_t \in X$ is a variable selected for branching, at step $t$ of the episode. The initial state $s_1 \in S$ corresponds to the situation after the execution of the fix-point at the root node.
A terminal node is reached either if all the variables are assigned ($\forall x \in X: |D_t(x)| = 1$), or if a 
failure is detected ($\exists  x \in X: |D_t(x)| = 0$). The variable selected for branching is obtained through a standard heuristic such as \textit{first-fail}.
\item[Set of actions] Given a state $s_t = (\mathcal{P}_t,x_t)$, an action $a_t$ corresponds to the selection of a value $v \in D(x_t)$ for branching at step $t$. 
Finding the most promising value to branch on is the problem addressed in this paper.
\item[Transition function] Given a state $s_t = (\mathcal{P}_t,x_t)$ and an action $a_t=v$, the transition function executes three successive operations. First, it assigns the value $v$ to the variable $x$ (i.e., $D(x_{t+1}) = v$). Second, it executes the fix-point on $\mathcal{P}_t$ in order to prune the domains (i.e., $\mathcal{P}_{t+1} = \mathsf{fixPoint}(P_{t})$).
Third, it selects the next variable to branch on (i.e., $x_{t+1} = \mathsf{nextVariable}(P_{t+1})$). This results
in a new state $s_{t+1} = (\mathcal{P}_{t+1},x_{t+1})$.
Integrating the propagation inside the transition is one important difference with Cappart et al.~\cite{cappart_combining_2020}.
\item[Reward function] The function is defined separately in Section~\ref{sec:reward}.
\end{description}

\begin{figure}[!ht]
\centering
    \begin{subfigure}[b]{0.468\textwidth}
        \centering
        \includegraphics[width=\textwidth]{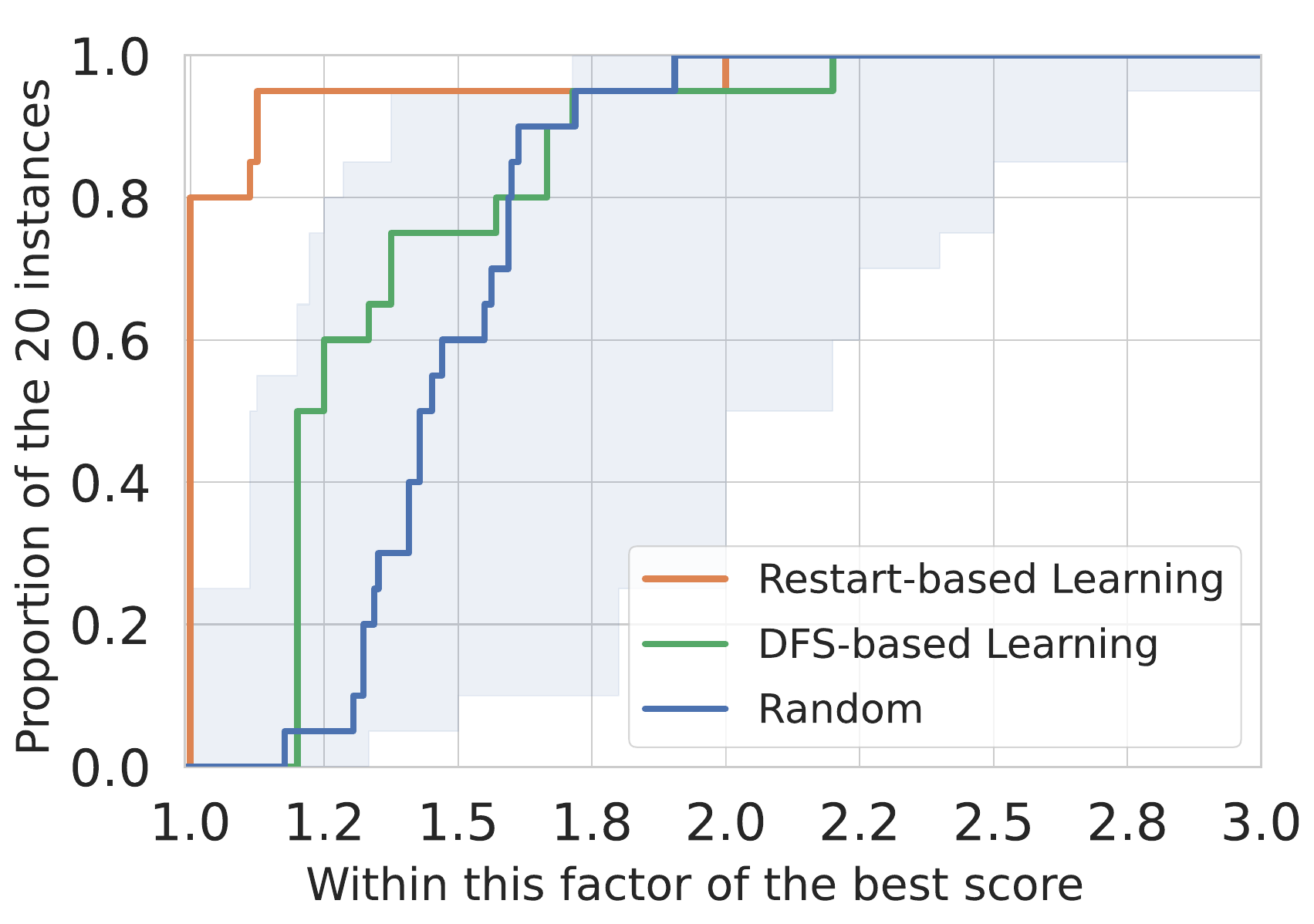}
        \caption{Score of the first solution obtained.}
        \label{fig:dfs_vs_cpnn_dive}
    \end{subfigure}
    \hfill
    \begin{subfigure}[b]{0.475\textwidth}
        \centering
        \includegraphics[width=\textwidth]{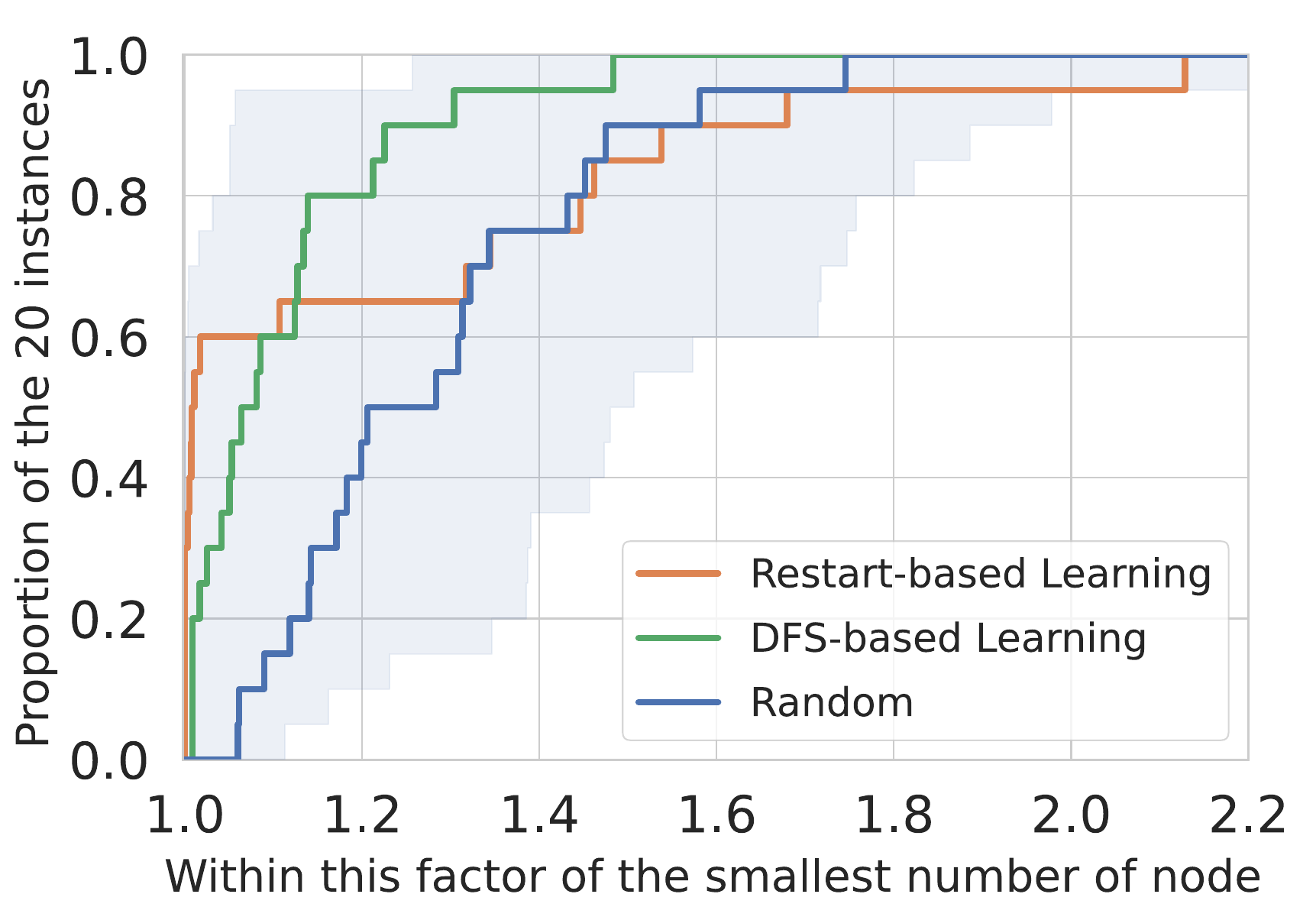}
        \caption{Number of node visited until optimality.}
        \label{fig:cfs_vs_cpnn_opt}
    \end{subfigure}
    \caption{Comparison of both training methods on maximum independent set (50 nodes). As a non-learned baseline, we added the performances of an agent performing only random decisions. Training is carried out on randomly generated Barabási-Albert graphs \cite{albert2002statistical}; we selected this type of distribution as the generated graphs are known to mimic human-made and natural organizations. The evaluation is performed on 20 other graphs following the same distribution.}
    \label{fig:dfs_vs_dives}
\end{figure}

Concerning the training, we opted for a \textit{double deep Q-learning} algorithm~\cite{van2016deep}, known to perform well for discrete action spaces. However, other RL algorithms could also be used. 
We compared our restart-based training procedure using a simple terminal reward based on the solution's score with the backtracking-based approach of Chalumeau et al.~\cite{chalumeau_seapearl_2021} using their reward at each step (penalty of $1$ for each explored node). We selected the \textit{maximum independent set problem} for this comparison with instances with 50 nodes. Results are presented using performance profiles~\cite{dolan2002benchmarking} in Figure~\ref{fig:dfs_vs_dives}. A detailed explanation of the experimental protocol is proposed in Section~\ref{sec:exp}. 

We evaluated both methods on two metrics matching the objective for which they were specifically trained. We look at the value of the solution obtained after a single dive (Figure~\ref{fig:dfs_vs_cpnn_dive}) in the tree search and the number of nodes visited to prove optimality using a depth-first search (Figure~\ref{fig:cfs_vs_cpnn_opt}). As expected, we observe that the agent trained with the \textit{restart-based learning} strategy allows good results regarding the optimality gap for the first solution found after a single dive. Remarkably, our method yields a comparable ability to prove optimality compared to Chalumeau et al.~\cite{chalumeau_seapearl_2021}, whose primary aim was specifically to solve the problem in the minimum number of nodes. This last result has to be mitigated as both RL-based methods lie in the range of the random strategy
(shaded blue area). 

Finally, as shown in Figure~\ref{fig:dfs_vs_cpnn_dive}, it is important to notice that the optimality gap returned by our method is still non-negligible at the first solution obtained. The complexity of a combinatorial problem lies mainly in closing this gap, which is why backtracking is required. Experiments with backtracking are proposed in Section \ref{sec:exp}.

\subsection{Propagation-Based Reward}
\label{sec:reward}
The definition of our reward must be aligned with our objective of \textit{finding quickly good solutions} for the combinatorial problem. Based on our training procedure, an intuitive function is to reward the agent proportionally to the solution quality found at the end of an episode. In case of an infeasible solution found, a penalty can be given. The main drawback of this rewarding scheme is that this information is only available at terminal nodes, and no reward is provided in branching nodes. This is related to the \textit{sparse reward} problem, which complicates the training process~\cite{trott2019keeping}. To address this challenge, one should find a way to give informative intermediate rewards along the solving process. To this end, we propose a new rewarding scheme based on the domain reduction of the objective variable (i.e., the variable that must be minimized or maximized). This reduction happens either thanks to the branching assignment or the application of the fix-point. There are two main components: (1) an \textit{intermediate reward} ($r^\mathsf{mid}$) collected at branching nodes, and (2) \textit{terminal reward} ($r^\mathsf{end}$) collected only at the end of an episode.

\begin{figure}[!ht]
    \centering
    \includegraphics[width=0.7\textwidth]{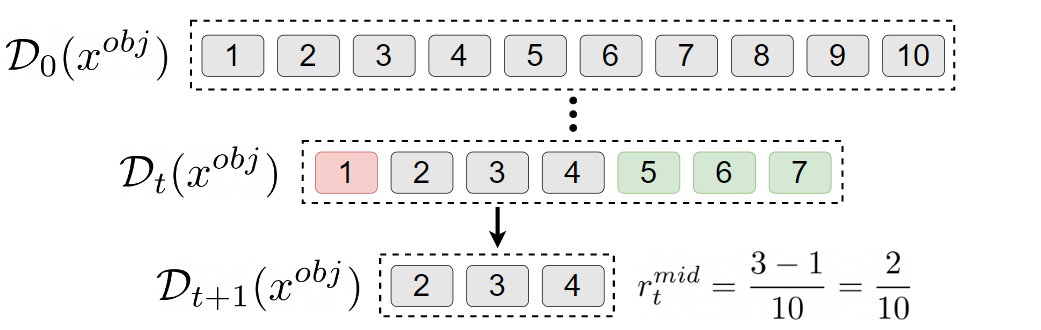}
    \caption{Intermediate reward when four values are pruned from the domain.}
    \label{fig:reward_schema}
\end{figure}

Assuming a minimization problem, the intermediate reward follows two principles: each domain reduction of the largest values of the domain is rewarded, 
and each domain reduction of the lowest values of the domain is penalized. It is important to note that following these principles does not guarantee the discovery of a good solution at the end of the branch. The rationale is to lead the agent to a situation where the minimum cost can be \textit{eventually} obtained while removing costly solutions.
It is formalized in Equations \eqref{eq:1} to \eqref{eq:3}, where $r^\mathsf{mid}_t$ is the reward obtained at step $t$, and is illustrated in Figure~\ref{fig:reward_schema}. 
As shown in Equation~\eqref{eq:terminal},
the terminal reward is set to -1 if the leaf node corresponds to an infeasible solution and 0 if it is feasible.
Finally, the total reward ($r^\mathsf{acc}$) accumulated during an episode of $T$ steps is the sum of all intermediate rewards with the final term, as proposed in Equation \eqref{eq:reward-fin}. 
\begin{align}
 r^\mathsf{ub}_t  &=  \# \Big\{ v \in D_t(x^{\mathsf{obj}}) ~ \Big|~ v \notin D_{t+1}(x^{\mathsf{obj}}) \land  v >  \max\big(D_t(x^{\mathsf{obj}})\big) \Big\} & \label{eq:1}\\
 r^\mathsf{lb}_t  &=  \# \Big\{ v \in D_t(x^{\mathsf{obj}}) ~ \Big|~ v \notin D_{t+1}(x^{\mathsf{obj}}) \land  v <  \min\big(D_t(x^{\mathsf{obj}})\big) \Big\} &\label{eq:2} \\
 r^\mathsf{mid}_t &= \frac{r^\mathsf{ub}_t - r^\mathsf{lb}_t}{\big|D_1(x^{\mathsf{obj}})\big|} & \label{eq:3} \\
 r^\mathsf{end}_t &= - 1 \mathsf{~if~unfeasible~solution~found~} (0 \mathsf{~otherwise})  & \label{eq:terminal} \\
 r^\mathsf{acc} &=  \Big( \sum_{t=1}^{T-1} r^\mathsf{mid}_t  \Big) +  r^\mathsf{end}_T & \label{eq:reward-fin}
\end{align}

\begin{figure}[!ht]
    \centering
    \begin{subfigure}[b]{0.32\textwidth}
        \centering
        \includegraphics[width=\textwidth]{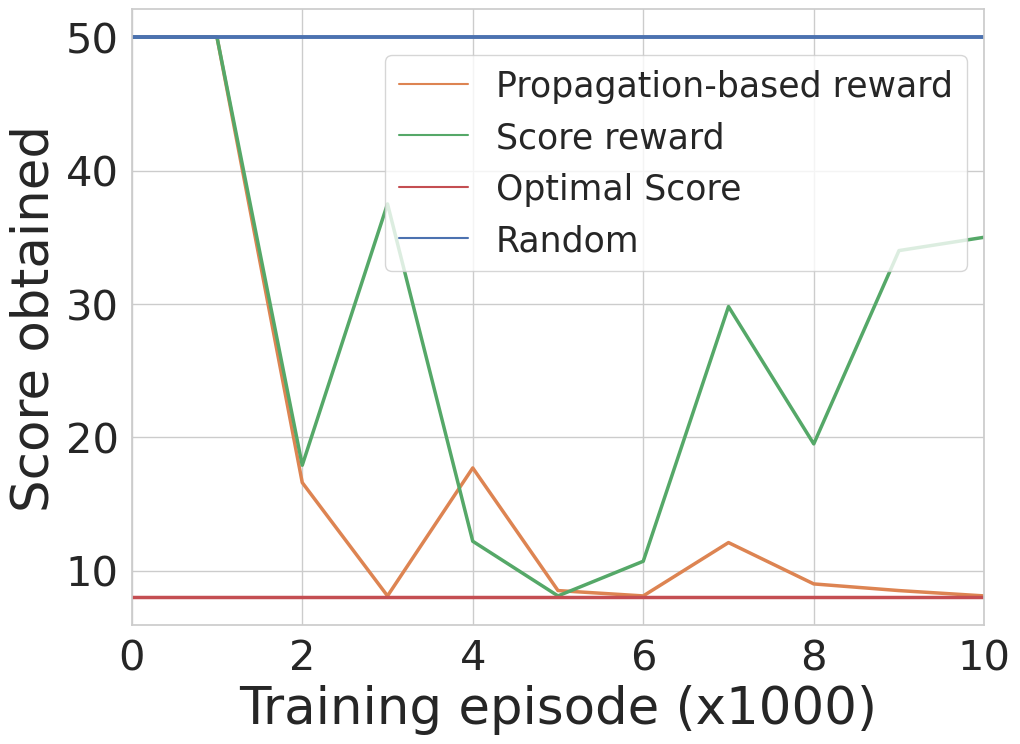}
        \caption{Graph coloring.}
        \label{fig:intermediate_rewards_gc}
    \end{subfigure}
    \begin{subfigure}[b]{0.32\textwidth}
        \centering
        \includegraphics[width=\textwidth]{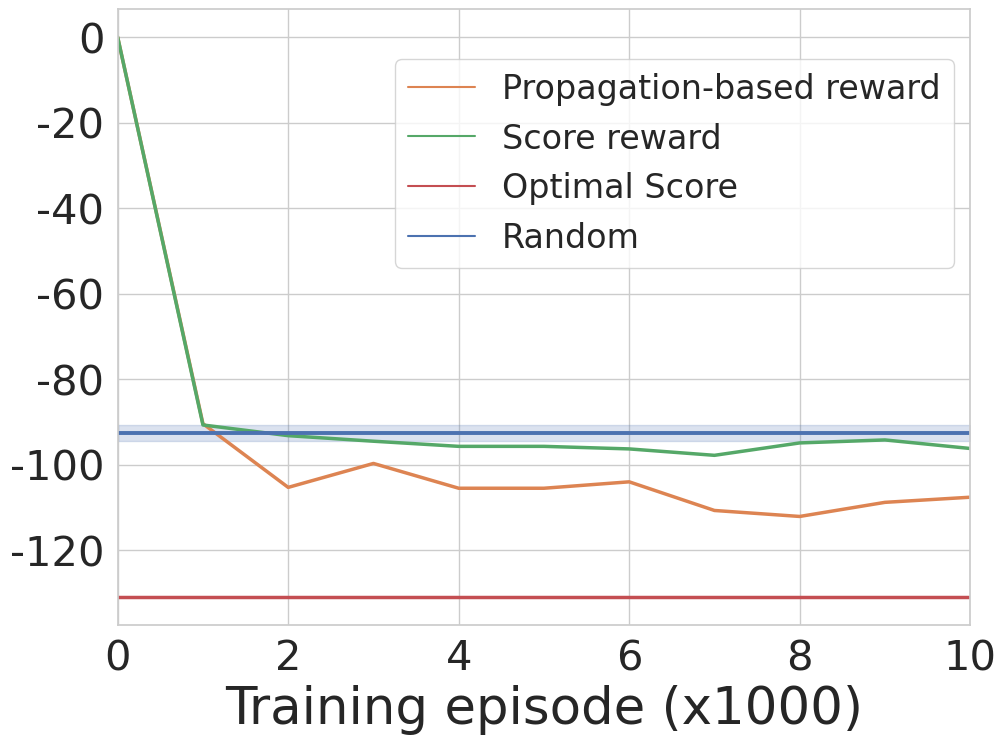}
        \caption{Maximum cut.}
        \label{fig:intermediate_rewards_mc}
    \end{subfigure}
    \begin{subfigure}[b]{0.31\textwidth}
        \centering
        \includegraphics[width=\textwidth]{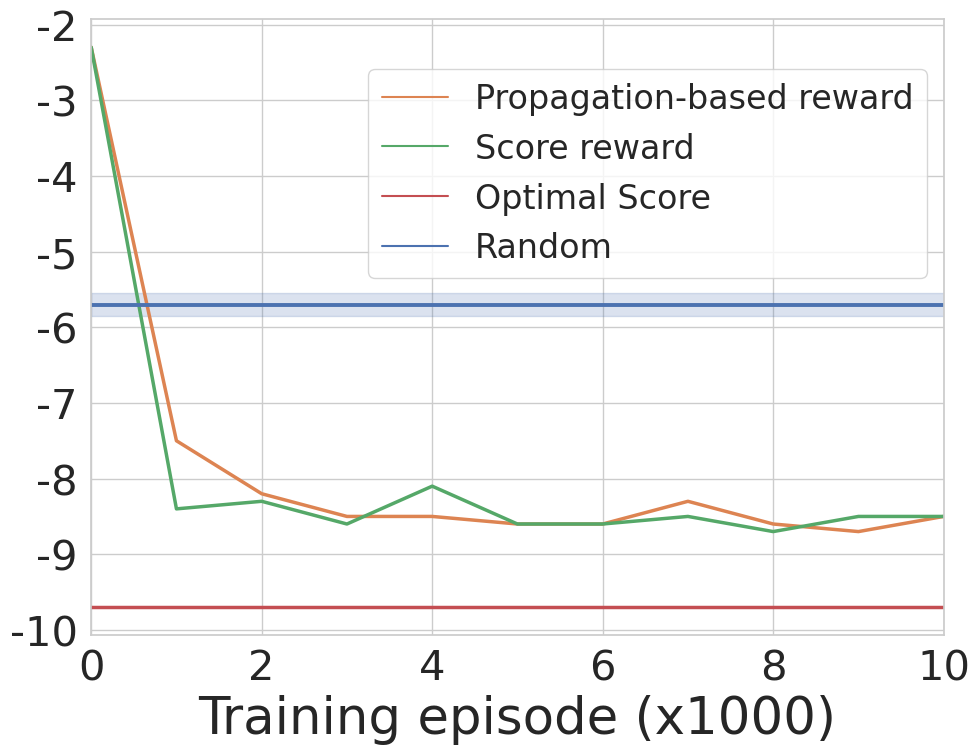}
        \caption{Maximum independent set.}
        \label{fig:intermediate_rewards_mis}
    \end{subfigure}
    \caption{Training curve for the two rewarding schemes, each validation step corresponds to performing a single dive in the search tree, the \textit{score} obtained refers to the quality of the solution found on the leaf node.}
    \label{fig:intermediate_rewards}
\end{figure}

An experimental analysis of this new reward scheme (\textit{propagation-based reward}) is carried out for the \textit{graph coloring}, \textit{maximum cut}, and \textit{maximum independent set} problems; we look at the quality of the solution found after a single dive in the search tree. As a baseline, we consider a reward (\textit{score reward}) that only gives a value at terminal nodes ($r^\mathsf{end}_T$) without an intermediate reward. Besides, we also consider the solutions returned by a random value-selection heuristic as a baseline. Figure~\ref{fig:intermediate_rewards} shows the evolution of the quality of the first solution returned ($y$-axis, averaged on 20 instances of the validation step) with the training time (number of episodes in the $x$-axis) using for training our restart-based search strategy defined in Section~\ref{subsec:rbs}. Instances are Barabási-Albert randomly generated graphs with 50 nodes. Except for the rewarding scheme, the other parts of the architecture are unchanged. We observe that the \textit{propagation-based reward} provides a more stable training (Figure~\ref{fig:intermediate_rewards_gc}) and can converge to a better model or, at least, to an equally good model as the terminal \textit{score reward} (Figures~\ref{fig:intermediate_rewards_mc} and \ref{fig:intermediate_rewards_mis}).

It should be noted that depending on the problem, the reward signal may remain sparse inside episodes even with our definition; this explains the discrepancy across the three class problems. Indeed, constraint propagation might take several steps to reach the objective variable, meaning that for related intermediate decisions, no value will be pruned from the domain of the objective variable. The \textit{graph coloring} problem is thus the problem for which taking these intermediate rewards is the most beneficial. Indeed, any previously unused color added will negatively impact the domain of the objective function, yielding an insightful negative reward. Conversely, branching on the  \textit{maximum independent set} problem does not consistently impact the objective function domain through the mechanism of constraint propagation, particularly at the beginning of the search. Our method yields no worse result than the usual reward signal in this setting. This worst-case scenario empirically validates the robustness of this reward.

\subsection{Heterogeneous Graph Neural Network Architecture}
\label{sec:archi}
\begin{figure}[!ht]
\centering
\includegraphics[width=\textwidth]{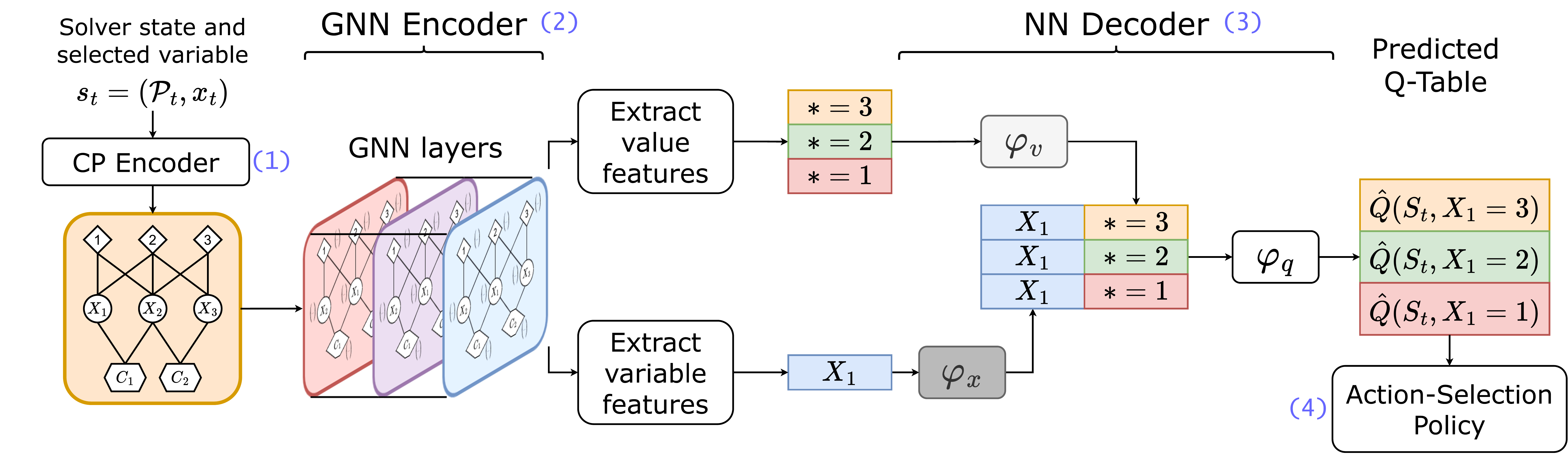}
\caption{High-level overview of the neural architecture designed.}
\label{fig:cpnn}
\end{figure}

An important part of the framework is the neural network architecture that
we designed to perform a prediction of the next value to branch on.
A high-level representation is proposed in Figure~\ref{fig:cpnn}.
Four steps are carried out: 
(1) a \textit{CP model encoder}, 
(2) a \textit{graph neural network encoder},
(3) a \textit{neural network decoder}, and
(4) an \textit{action-selection policy}.
They are detailed in the next subsections.

\subsubsection*{Step 1: CP Model Encoder}

\begin{figure}[htbp]
    \centering
    \includegraphics[width=1\textwidth]{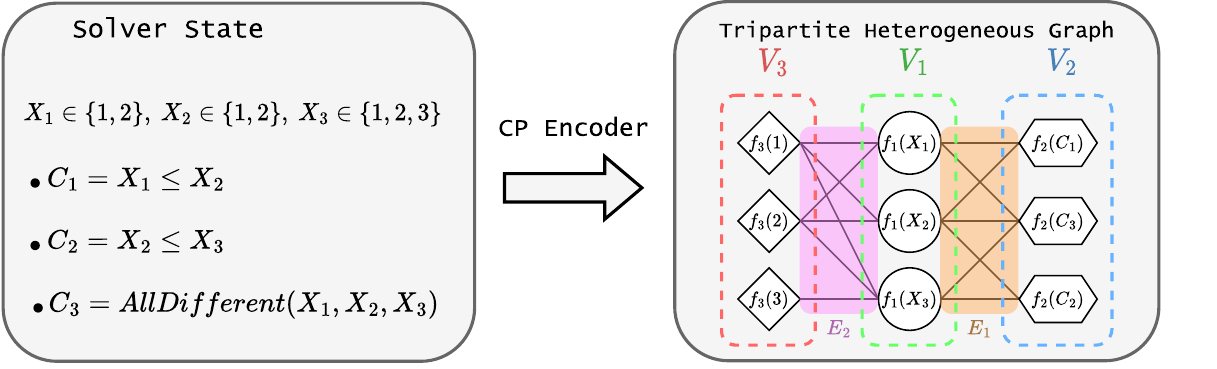}
    \caption{Representation computed by the CP encoder on a simple example.}
    \label{fig:cp_encoder}
\end{figure}

The core idea is to learn for any CP model given as input, unlike Cappart et al.~\cite{cappart_combining_2020}, who require a specific encoding for each combinatorial problem.
This has been achieved for mixed-integer programs thanks to a bipartite graph representation~\cite{gasse_exact_2019}
and by Chalumeau et al.~\cite{chalumeau_seapearl_2021} for CP models thanks to a tripartite graph.
This last work does not leverage any feature related to the variables, values, or constraints.
We built upon this last approach by adding such features.
Specifically, let $\mathcal{P} = \langle X, D(X), C, O \rangle$ be the combinatorial problem we want to encode.
The idea consists in building 
a simple undirected graph $\mathcal{G}(V_1,V_2,V_3,f_1,f_2,f_3,E_1, E_2)$ encoding all the information of $\mathcal{P}_t$
from a state $s_t = (\mathcal{P}_t,x_t)$.
In this representation, $V_1$, $V_2$, and $V_3$ are three sets of vertices, $f_1$, $f_2$, and $f_3$ are
three sets of feature vectors, and $E_1$ with $E_2$ are two distinct sets of edges.
This yields a graph with three types of nodes decorated with features. The first part of the encoding we propose is as follows:
(1) each variable, constraint, and value corresponds to a specific type of node ($V_1 =X$, $V_2 = C$, and $V_3 = D$),
(2) each time a variable $x \in V_1$ is involved in a constraint $c \in V_2$, an  edge $(x,c) \in  E_1$ is added between both nodes,
(3) each time a value $v \in V_3$ is in the domain of a variable $x \in V_1 $, an edge $(v,x) \in  E_2$ is added between both nodes.
This gives a tripartite graph representation of a CP model generically.
This is illustrated in Figure~\ref{fig:cp_encoder}.
The second part of the encoding is to add features to each node. 
Intuitively, the features will provide meaningful information and thus improve the quality of the model.
The features we considered are proposed below.
We note that we can easily extend this encoding by integrating new features. 
\begin{enumerate}
\item \textit{Features attached to variables ($f_1$)}: the current domain size, the initial domain size, 
a binary indication if the variable is already assigned, and a binary indication if the variable corresponds to the objective.
\item  \textit{Features attached to constraints ($f_2$)}: the constraint type (one-hot encoding), and a binary indication
if the constraint propagation has reduced domains.
\item  \textit{Features attached to values ($f_3$)}: its numerical value.
\end{enumerate}

\subsubsection*{Step 2: Graph Neural Network Encoder}

Once the CP model has been encoded as a graph, the next step
is to embed this representation as a latent vector of features for each node
of the graph (see Section \ref{sec:gnn}).
We propose to carry out this operation with a graph neural network.
Unlike the standard prediction scheme presented in Equation \eqref{eq:gnn},
our graph has three types of nodes. For this reason,
we opted for a \textit{heterogeneous} architecture.
Concretely, a specific convolution is carried out for each node type. The architecture is detailed in Equations \eqref{eq:gnn_hetero_1} to \eqref{eq:gnn_hetero_3}, where $\bigoplus$
is the \textit{sum-pooling} or \textit{mean-pooling} aggregation,
operator $(.\|.)$ is a concatenation of vectors, 
$N_x(n)$ is the set of neighbouring nodes of $n$ from $V_1$ (variable),
$N_c(n)$ is the set of neighbouring nodes of $n$ from $V_2$ (constraint),
$N_v(n)$ is the set of neighbouring nodes of $n$ from $V_3$ (value),
$\theta_{1,\dots,10}^k$ 
are weight matrices at layer $k$, 
and $g$ is the \textsf{leakyReLU} activation function~\cite{maas2013rectifier}.
Another difference with the canonical GNN equation
is the integration of \textit{skip connections} ($h_x^{0}$, $h_c^{0}$, and $h_c^{0}$)
allowing to keep at each layer information from the input features. 
This technique is ubiquitous in deep convolutional networks 
such as in \textit{ResNet}~\cite{he2016deep}.
Finally, the initial embedding are initialized as follows:
$h_x^{0} = \theta_{11} f_1$, $h_c^{0} = \theta_{12} f_2$, and $h_v^{0} = \theta_{13} f_3$,
where $\theta_{11,\dots,13}$ are new weight matrices.
\begin{align}
\label{eq:gnn_hetero_1}
    h_{x}^{k+1} &= g \Big(\theta_1^k h_x^{0}  ~ \big\| ~ \theta_2^k h_x^{k}  ~ \big\| ~ (\bigoplus_{c \in N_c(x)} \theta_3^k h_c^{k})    ~ \big\| ~ (\bigoplus_{v \in N_v(x)} \theta_4^k h_v^{k}) \Big) & \forall{x \in V_1}\\
\label{eq:gnn_hetero_2}
    h_{c}^{k+1} &= g \Big(\theta_5^k h_c^{0}  ~ \big\| ~  \theta_6^k h_c^{k}  ~ \big\| ~ (\bigoplus_{x \in N_x(c)} \theta_7^k h_x^{k})  \Big) &  \forall{c \in V_2} \\
\label{eq:gnn_hetero_3}
    h_{v}^{k+1} &= g \Big(\theta_8^k h_v^{0}  ~ \big\| ~  \theta_9^k h_v^{k}  ~ \big\| ~ (\bigoplus_{x \in N_x(v)} \theta_{10}^k h_x^{k})  \Big) & \forall{v \in V_3}
\end{align}

\subsubsection*{Step 3: Neural Network Decoder}

At this step, a $d$-dimensional tensor is obtained for each graph node.
Let $x \in V_1$ be the node representing the current variable selected for branching, and $V_x \subseteq V_3$ the subset of nodes representing the 
values available for $x$ (i.e., the values that are in the domain of the variable).
The goal of the \textit{decoder} is to predict a \textit{Q-value} (see Section \ref{sec:RL}) for each $v \in V_x$.
The computation is formalized in Equation \eqref{eq:decoder},
where $h_{x}^{K}$ and $h_{v}^{K}$ are the node embedding of variable $x$ and 
value $v$, respectively, after $K$ iterations of the GNN architecture.
The functions $\varphi_x: 
\mathbb{R}^{d} \to 
\mathbb{R}^l, \varphi_v: 
\mathbb{R}^{d} \to 
\mathbb{R}^l, \varphi_q: 
\mathbb{R}^{2l} \to 
\mathbb{R}$ are fully-connected neural networks.
Such a $Q$-value must be computed for each value $v \in V_x$. 
It is internally done thanks to matrix operations, allowing a more
efficient computation.
\begin{equation}
\label{eq:decoder}
    \hat{Q}(h_{x}^{K}, h_{v}^{K}) = \varphi_q \Big( \varphi_x(h_{x}^{K}) ~ \big\| ~ \varphi_v( h_{v}^{K}) \Big) ~ ~ \forall v \in V_x
\end{equation}

\subsubsection*{Step 4: Action-Selection Policy}

Once all the $Q$-values have been computed for the current variable, the policy is defined by an \textit{explorer} that can decide to exploit the approximated $Q$-values by greedily choosing the best action as shown in Equation \eqref{eq:pred} or decide to select unpromising action associated with a lower $Q$-value (for example, by selecting a random action with probability $\epsilon$). This behavior derives from the trade-off between exploitation and exploration, which is necessary for early learning when the estimates of Q-values are poor, and when only a few states have been visited. Once trained, the $Q$-values should represent the branching choice leading to the best decision according to the reward of Equation \eqref{eq:reward-fin}.
\begin{equation}
\label{eq:pred}
\pi(v | x) = \mathsf{argmax}_{v \in V_x} \hat{Q}(h_{x}^{K}, h_{v}^{K})
\end{equation}

Assembling  all the pieces, this architecture gives a generic approach to obtaining a data-driven value-selection heuristic inside a CP solver. Concerning the search strategy used for evaluation (which is different from the \textit{restart-based} one used for training), we propose to embed our predictions inside an \textit{iterative limited discrepancy search} (ILDS) \cite{LDS}. This strategy is commonly used when we are confident in the quality of the  heuristic. The core idea is to restrict the number of branching choices deviating from the heuristic (i.e., a \textit{discrepancy}). By doing so, the search will explore a subset of solutions expected to be good while giving a chance to reconsider the value-heuristic selection which is nevertheless prone to errors. This mechanism is enriched with a procedure that iteratively increases the number of discrepancies allowed once a level has been explored.

\section{Experiments}
\label{sec:exp}
The goal of this section is to evaluate the quality of the learned value-selection heuristic and the efficiency of the approach. Three combinatorial optimization problems are considered: \textit{graph coloring} (\textsf{COL}),  \textit{maximum independent set} (\textsf{MIS}), and \textit{maximum cut} (\textsf{MAXCUT}).

\subsection{Experimental Protocol}

Three configurations for the distribution of the problems generated are proposed for each problem: \textit{small} (20 to 30 nodes), \textit{medium} (40 to 50 nodes), and \textit{large} (80 to 100 nodes) instances, except for $\textsf{MAXCUT}$ which was already challenging for the \textit{medium} size.  Training is carried out on randomly generated Barabási-Albert graph \cite{albert2002statistical} with a density factor varying between 4 and 15 according to the size of the instances. A specific model is trained for each configuration of each combinatorial problem. The training is done using randomly generated instances.  Evaluation is then performed on 20 new graphs following the same distributions. The models are trained on an Nvidia Tesla V100 32Go GPU until convergence. It took up to 72 hours of training time for the most difficult cases (\textit{graph coloring} with 80 nodes) and less than 1 hour for the simplest cases (\textit{graph coloring} with 20 nodes). Each operation of the CP solver during training and evaluation is carried out on a CPU Intel Xeon Silver 4116 at 2.10GHz. The approach has been implemented in \textit{Julia} and is integrated into the solver \textit{Seapearl}. The implementation is available  on GitHub with BSD 3-Clause licence\footnote{https://github.com/corail-research/SeaPearl.jl}.

We compared our approach (\textsf{Learned, ILDS}) with two other generic value selection heuristics: \textit{impact-based search} (\textsf{Impact}) \cite{refalo2004impact} and \textit{activity-based search} (\textsf{Activity}) \cite{michel2012activity}. The standard \textit{minDomain} heuristic is used for the variable selection. Comparisons with Chalumeau et al.~\cite{chalumeau_seapearl_2021} have been provided in Section \ref{subsec:rbs}. As it has been highlighted that this approach is not suited to find good solutions quickly, it is not included again in the next experiments.
Each approach is evaluated with a fixed node budget depending on the parameters of the distribution used to generate the problems. For our approach, the performance obtained after the first dive in the tree search is also monitored (\textsf{Learned, \nth{1} dive}). 
As \textsf{Impact} and \textsf{Activity} are online learning methods, they perform similarly to a random selection at the beginning of the search. For this reason, the performance
obtained after the first dive in the tree search with such methods is omitted. 
Finally, we also included a comparison with a random selection using DFS with the same node budget (\textsf{Random}).
Finally, the optimal cost (\textsf{OPT}) has been obtained with an exact approach without any restriction on the budget.

\subsection{Quantitative Results}

Table \ref{table:benchmark_score} summarizes the main results of our approach. As a general comment, our approach can find solutions of superior quality given a node budget or find the optimal solution by exploring fewer nodes than the baselines. 
Interestingly, our approach (\textsf{Learned, ILDS}) can learn a branching strategy giving high-quality solutions, even without backtracking (\textsf{\nth{1} dive}). For instance, a single dive for \textit{maximum cut} with 50 nodes yields almost instantly a solution with an optimality gap of 0.16, whereas a depth-first search with a random selection (\textsf{Random, DFS}) required 19 seconds and roughly 53,000 nodes explored to find a solution with the same gap. Within this same budget, (\textsf{Learned, ILDS}) significantly improves the solution and achieves an optimality gap of 0.09. 
It is worth highlighting that (\textsf{Learned, ILDS}) took 130 seconds to explore 38,744 nodes and has, thereby, an exploration rate slower than the other methods. This significantly increased execution time is mainly because calling the graph neural network architecture (Section \ref{sec:archi}) at each tree search node is much more computationally expensive than calling a simple heuristic. This difficulty is further discussed in Section \ref{dis_lim}.

\begin{table}[!ht] 
\centering
\caption{Results for the three problems given a fixed node budget.
The average result (rounded) on the 20 test instances is reported for each configuration.
\textit{Gap} indicates the optimality gap,
\textit{Node} gives the number of nodes explored before finding the best solution within the budget,
and \textit{Time} gives the time (seconds) before finding this solution. }
\label{table:benchmark_score}

\resizebox{\textwidth}{!}{%
\begin{tabular}{lrr|c|rrr|rrr|rrr|rrr|c}                  
& &   & \multicolumn{4}{c|}{\textsf{Learned}} &\multicolumn{3}{c|}{\textsf{Activity-Based}} & \multicolumn{3}{c|}{\textsf{Impact-Based}} & \multicolumn{3}{c|}{\textsf{Random}}&      \\
& & &  \multicolumn{1}{c}{\textsf{\nth{1} dive}} & \multicolumn{3}{c|}{\textsf{ILDS}} &\multicolumn{3}{c|}{\textsf{DFS}} &\multicolumn{3}{c|}{\textsf{DFS}} & \multicolumn{3}{c|}{\textsf{DFS}}&      \\ 
& Size &  \textsf{OPT} & Gap & Gap & Node & Time & Gap & Node & Time & Gap & Node & Time & Gap & Node & Time & Budget\\

\hline

\multirow{3}{*}{\textsf{COL}} 

& 20 &  5.05 & 0.06&
0 & 27 & < 1 & 0 & 378 & < 1 & 0 & 374 & < 1 & 0 & 378 & < 1 & $10^3$ \\

& 40  & 7.90& 0.08
& 0 &104 & < 1 & 0 &1,664 & < 1& 0 & 1732& < 1& 0& 1735& < 1&$10^4$ \\

& 80  & 8.75& 0.06
& 0 & 120 & 1 & 0 & 7,051 & 2 & 0 & 7,057  & 2& 0 & 7,211 & 2& $10^5$  \\ 

\hline

\multirow{3}{*}{\textsf{MIS}} 

& 30 &  9.90 & 0.08
& 0 & 88 & < 1 & 0 & 215& < 1 & 0& 297 & < 1 & 0 & 293 & < 1&  $10^3$\\

& 50 &  15.00& 0.09
& 0& 539& 1& 0&  5,807& 1& 0& 7,474& 1& 0& 8,942& 1& $10^4$ \\

& 100 &  21.70&0.20
&0.02 &28,392 &253 &0.09 & 35,536& 7& 0.10& 38,154&8 & 0.10 & 41,774& 9&$10^5$   \\ 

\hline

\multirow{2}{*}{\textsf{MAXCUT}}   

& 20 &  46.70& 0.15
& 0.03 & 3,714 & 5 & 0.04 & 4,635 & 1 & 0.03& 5,959 & 2& 0.04 & 4877& 1 & $10^4$   \\
                                
& 50 &  222.00 & 0.16
&0.09 & 38,744 & 130 & 0.17 & 44,664 & 14 & 0.17 & 47,970& 17 & 0.17& 53,110& 19 &$10^5$   \\ 
\end{tabular}
}
\end{table}

\begin{figure}[!ht]
\centering
    \begin{subfigure}[b]{0.48\textwidth}
        \centering
        \includegraphics[width=\textwidth]{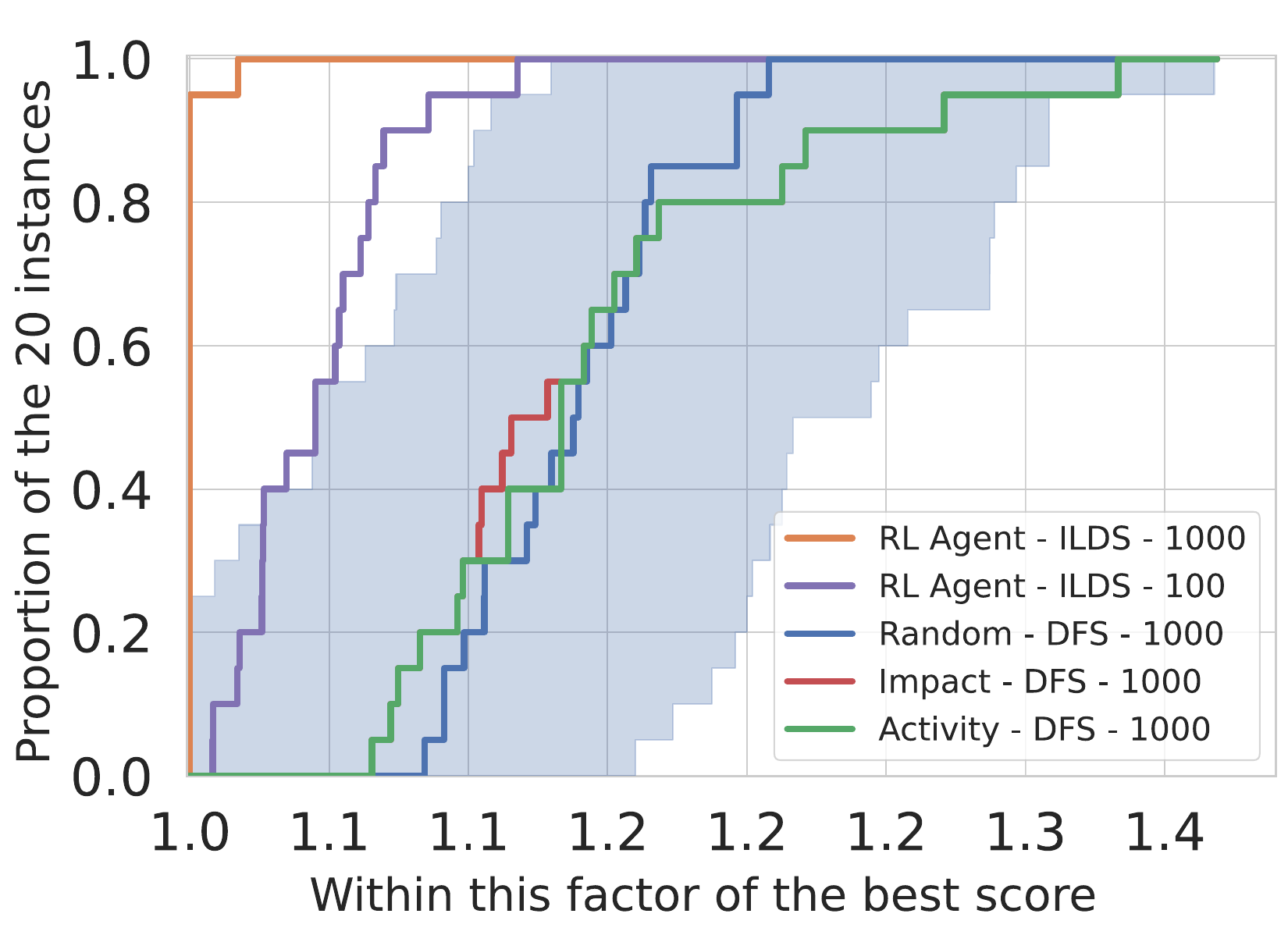}
        \caption{Maximum cut with 50 nodes.}
        \label{fig:performance_plot_MC}
    \end{subfigure}
    \begin{subfigure}[b]{0.48\textwidth}
        \centering
        \includegraphics[width=\textwidth]{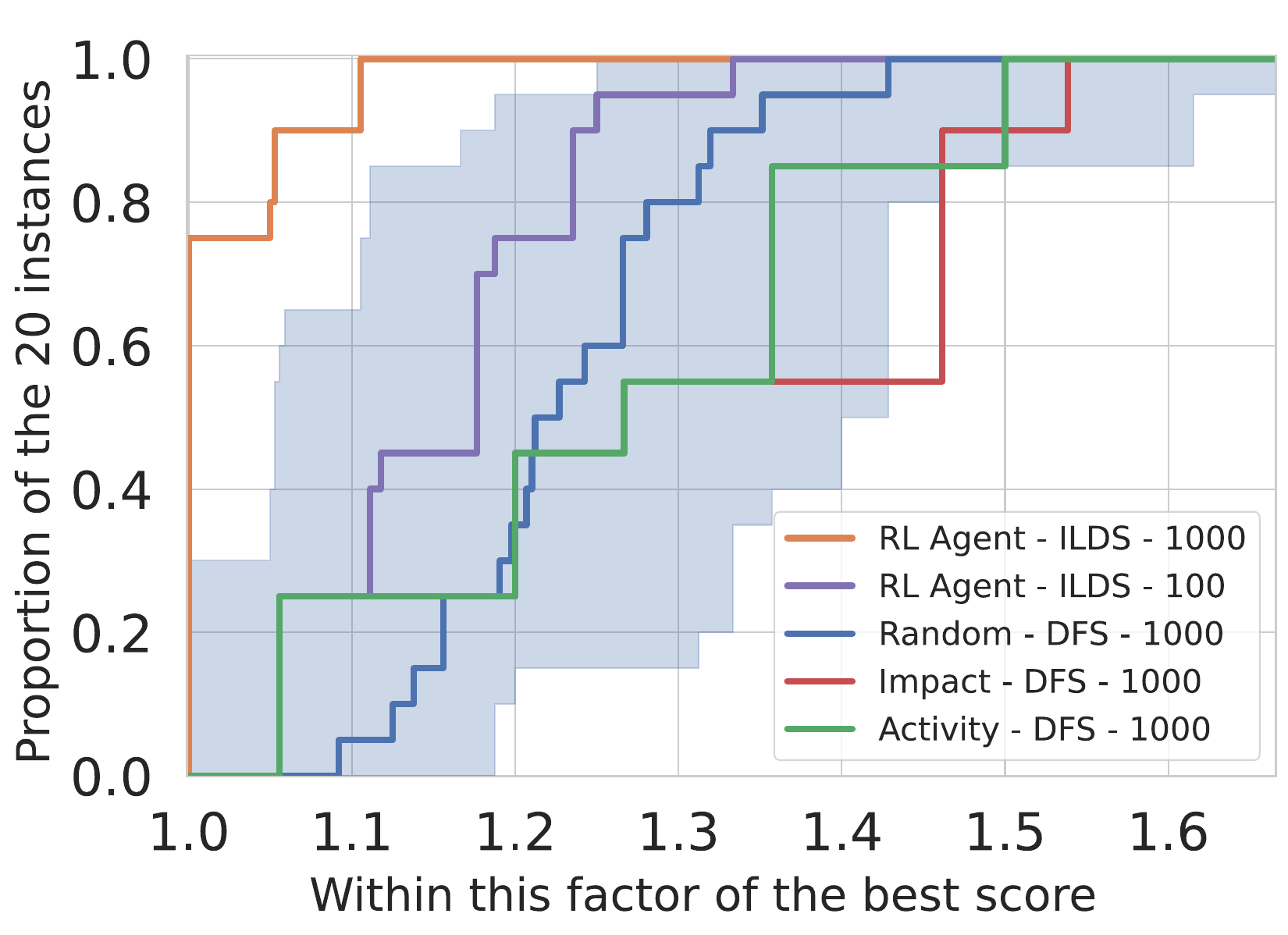}
        \caption{Maximum independent set with 100 nodes.}
        \label{fig:performance_plot_MIS}
    \end{subfigure}
    \caption{Best solutions found within a restricted node budget on largest instances for the three problems considered. We set a small budget to evaluate the ability of each approach to \textit{find quickly a good solution}, which is the objective aimed by this work.
    The performance profile ratio is computed using the optimal solution as a reference. 
    Within the same maximal number of nodes visited (1000), we observe that (\textsf{Learned, ILDS}) dominate all the other methods. 
    Besides, we still perform better than the baselines when restricting ten times the budget for \textsf{ILDS-Learned}.}
    \label{fig:performance_plot}
\end{figure}

Concerning \textsf{Activity} and \textsf{Impact} heuristics, they yield no improvement on  \textit{graph coloring} compared to a random strategy. This can be explained by the fact that
this class of problem has many possible combinations of variables and values for branching. This requires a significantly larger number of explored nodes to initialize these two heuristics efficiently. 
For the two other problems, characterized by a binary domain for the values to branch on, \textsf{Activity} and \textsf{Impact} provide significantly better results than the random strategy, which is the expected behavior. In all the tested situations, 
(\textsf{Learned, ILDS}) provides the best optimality gap within the node budget.
Additional results are proposed in Figure~\ref{fig:performance_plot} using performance profiles~\cite{dolan2002benchmarking} for the two hardest situations (100 for \textit{maximum independent set}, and 50 for \textit{maximum cut}) given a node budget of 100 or 1000 nodes. 

\subsection{Discussions and Opportunities of Further Research}
\label{dis_lim}
The previous experiments showcased the promise of this framework to quickly find good solutions towards a generic value-selection heuristic inside a CP solver. There are nonetheless open challenges 
that must be considered for practical use. Four of them are discussed.

\subsubsection*{Challenge 1: Scalability of the Representation}

Our approach faces a double penalty regarding its \textit{scaling} capability: as the problem grows larger,  the tripartite representation increases significantly in size, which results in a longer computation time required to make one branching decision. This impacts both training and evaluation. Additionally, the number of nodes (and, therefore, decisions to be made) in the search tree grows exponentially with the problem size, exacerbating the aforementioned phenomenon. Consequently, our approach is penalized twice due to the exponential behavior of combinatorial problems.
As a concrete example, \textit{graph coloring} instances with 80 nodes require 72 hours of training on a GPU, while only 1 hour is required for the smallest instances.
An interesting research direction to mitigate this difficulty is to build a mechanism to  compact the representation, for instance, thanks to network pruning tools \cite{yu2022combinatorial} or with  \textit{transfer learning}. 
Another idea is to call the model only in a few nodes, in a similar fashion as Cappart et al.~\cite{cappart2022improving} did for \textit{decision-diagram-based branch-and-bound}~\cite{bergman2016discrete}.
On a lower level of computation, standard constraint programming solvers perform sequential decisions and  are therefore optimized for CPU architecture. Concerning the training, 
it is carried out on a GPU. In the current implementation, each branching decision requires loading the entire tripartite graph on the Video RAM, which is inefficient. 
We believe much work could be done to optimize this CPU/GPU architecture, for instance by delegating other operations
on the GPUs, such as the propagation of few constraints~\cite{campeotto2014exploring,tardivo2022constraints}. 
    
\subsubsection*{Challenge 2: Tackling Highly Constrained Problems}

The experiments proposed in the paper considered combinatorial problems where the difficulty lay in finding the \textit{best} solution. Still, it was easy to find a \textit{feasible} solution, even of poor quality. We empirically observed that the learning performance largely depends on the abundance of feasible solutions in the search space. This is explained by the definition of the reward, which is based on the propagation occurring on the objective variable (see $r_t^{\mathrm{mid}}$ in Section \ref{sec:reward}).
However, when feasible solutions are not easily obtained, such as in highly constrained problems, the reward signal becomes less informative. Addressing such combinatorial problems remains an open challenge. 
We believe an extension of the reward signal can address this in order to handle other situations.

\subsubsection*{Challenge 3: Learning a Combined Variable/Value Heuristic}

Although this work proposes to learn a value-selection heuristic, 
learning how to branch on variables has already been considered 
in the literature~\cite{song_learning_2022}.
An interesting research direction is to adapt this architecture to learn 
a variable-selection and a value-selection heuristic in a unified way. 
A possible direction is to consider a model with a double-head decoder, the first for selecting the variable and the second for selecting the value. 
On the training aspect, two reinforcement learning agents could be trained, with an
the incentive to cooperate with the information sharing \cite{sunehag2017valuedecomposition}.

\subsubsection*{Challenge 4: Proving the Optimality of a Solution}
The goal pursued in this paper is to find the best solution as quickly as possible.
Another direction is to guide the search to speed-up the optimality proof. It is what has been proposed by Chalumeau et al.~\cite{chalumeau_seapearl_2021}.
In practice, finding good solutions and proving optimality 
are complementary aspects inside a constraint programming solver and should be both considered.
Possible directions to do so could be to  redefine the reward function appropriately or
to revise the definition of an episode, as proposed by Scavuzzo et al. with TreeMDPs~\cite{treeMDP}.

\section{Conclusion}
\label{sec:conclusion}
The efficiency of constraint programming solvers is partially due to the branching heuristics used to guide the search. In practice, value-selection heuristics are often designed thanks to problem-specific expert knowledge, often out of reach for non-practitioners. In this paper, we proposed  a method based on reinforcement learning for obtaining such a heuristic, thanks to historical data, characterized by problem instances following the same distribution of the one that must be solved. This has been achieved thanks to a restart-based training procedure, a non-sparse reward signal, and a heterogeneous graph neural network architecture. Experiments on three combinatorial optimization problems show that the framework can find better solutions close to optimality in fewer nodes visited than other generic baselines. Several limitations and challenges (e.g., tractability for larger or real-world instances, transfer learning, sparsity of the reward signal) have been identified, and addressing them is part of future work. We also plan to consider other combinatorial problems, such as the ones proposed in XCSP3 competitions~\cite{audemard2022proceedings}.



\bibliography{main}

\begin{thebibliography}{10}

\bibitem{SpinningUp2018}
Joshua Achiam.
\newblock Spinning up as a deep {RL} researcher, Oct 2018.
\newblock URL:
  \url{spinningup.openai.com/en/latest/spinningup/spinningup.html}.

\bibitem{albert2002statistical}
R{\'e}ka Albert and Albert-L{\'a}szl{\'o} Barab{\'a}si.
\newblock Statistical mechanics of complex networks.
\newblock {\em Reviews of modern physics}, 74(1):47, 2002.

\bibitem{audemard2022proceedings}
Gilles Audemard, Christophe Lecoutre, and Emmanuel Lonca.
\newblock Proceedings of the 2022 {XCSP3} competition.
\newblock {\em arXiv preprint arXiv:2209.00917}, 2022.

\bibitem{bello_neural_2016}
Irwan Bello, Hieu Pham, Quoc~V. Le, Mohammad Norouzi, and Samy Bengio.
\newblock Neural combinatorial optimization with reinforcement learning.
\newblock {\em arXiv preprint arXiv:1611.09940}, 2016.

\bibitem{bengio_machine_2020}
Yoshua Bengio, Andrea Lodi, and Antoine Prouvost.
\newblock Machine learning for combinatorial optimization: a methodological
  tour d’horizon.
\newblock {\em European Journal of Operational Research}, 290(2):405--421,
  2021.

\bibitem{bergman2016discrete}
David Bergman, Andre~A. Cire, Willem-Jan van Hoeve, and John~N. Hooker.
\newblock Discrete optimization with decision diagrams.
\newblock {\em INFORMS Journal on Computing}, 28(1):47--66, 2016.

\bibitem{bonami2018learning}
Pierre Bonami, Andrea Lodi, and Giulia Zarpellon.
\newblock Learning a classification of mixed-integer quadratic programming
  problems.
\newblock In {\em International Conference on the Integration of Constraint
  Programming, Artificial Intelligence, and Operations Research}, pages
  595--604. Springer, 2018.

\bibitem{campeotto2014exploring}
Federico Campeotto, Alessandro Dal~Palu, Agostino Dovier, Ferdinando Fioretto,
  and Enrico Pontelli.
\newblock Exploring the use of {GPU}s in constraint solving.
\newblock In {\em Practical Aspects of Declarative Languages: 16th
  International Symposium, PADL 2014, San Diego, CA, USA, January 20-21, 2014.
  Proceedings 16}, pages 152--167. Springer, 2014.

\bibitem{cappart2022improving}
Quentin Cappart, David Bergman, Louis-Martin Rousseau, Isabeau
  Pr{\'e}mont-Schwarz, and Augustin Parjadis.
\newblock Improving variable orderings of approximate decision diagrams using
  reinforcement learning.
\newblock {\em INFORMS Journal on Computing}, 2022.

\bibitem{cappart_GNN}
Quentin Cappart, Didier Chételat, Elias~B. Khalil, Andrea Lodi, Christopher
  Morris, and Petar Velickovic.
\newblock Combinatorial optimization and reasoning with graph neural networks.
\newblock {\em Journal of Machine Learning Research}, 24(130):1--61, 2023.
\newblock URL: \url{http://jmlr.org/papers/v24/21-0449.html}.

\bibitem{cappart_combining_2020}
Quentin Cappart, Thierry Moisan, Louis-Martin Rousseau, Isabeau
  Pr{\'e}mont-Schwarz, and Andre~A. Cire.
\newblock Combining reinforcement learning and constraint programming for
  combinatorial optimization.
\newblock In {\em Proceedings of the AAAI Conference on Artificial
  Intelligence}, volume~35, pages 3677--3687, 2021.

\bibitem{chalumeau_seapearl_2021}
F{\'e}lix Chalumeau, Ilan Coulon, Quentin Cappart, and Louis-Martin Rousseau.
\newblock Seapearl: A constraint programming solver guided by reinforcement
  learning.
\newblock In {\em International Conference on Integration of Constraint
  Programming, Artificial Intelligence, and Operations Research}, pages
  392--409. Springer, 2021.

\bibitem{chu2015learning}
Geoffrey Chu and Peter~J. Stuckey.
\newblock Learning value heuristics for constraint programming.
\newblock In {\em International Conference on Integration of Artificial
  Intelligence and Operations Research Techniques in Constraint Programming for
  Combinatorial Optimization Problems 2015}, pages 108--123. Springer, 2015.

\bibitem{dolan2002benchmarking}
Elizabeth~D. Dolan and Jorge~J. Mor{\'e}.
\newblock Benchmarking optimization software with performance profiles.
\newblock {\em Mathematical programming}, 91(2):201--213, 2002.

\bibitem{doolaard2022online}
Floris Doolaard and Neil Yorke-Smith.
\newblock Online learning of variable ordering heuristics for constraint
  optimisation problems.
\newblock {\em Annals of Mathematics and Artificial Intelligence}, pages 1--30,
  2022.

\bibitem{gasse_exact_2019}
Maxime Gasse, Didier Ch{\'e}telat, Nicola Ferroni, Laurent Charlin, and Andrea
  Lodi.
\newblock Exact combinatorial optimization with graph convolutional neural
  networks.
\newblock {\em Advances in Neural Information Processing Systems}, 32, 2019.

\bibitem{glorot2011deep}
Xavier Glorot, Antoine Bordes, and Yoshua Bengio.
\newblock Deep sparse rectifier neural networks.
\newblock In {\em Proceedings of the fourteenth international conference on
  artificial intelligence and statistics}, pages 315--323. JMLR Workshop and
  Conference Proceedings, 2011.

\bibitem{haarnoja2018soft}
Tuomas Haarnoja, Aurick Zhou, Pieter Abbeel, and Sergey Levine.
\newblock Soft actor-critic: Off-policy maximum entropy deep reinforcement
  learning with a stochastic actor.
\newblock In {\em International conference on machine learning}, pages
  1861--1870. PMLR, 2018.

\bibitem{LDS}
William~D. Harvey and Matthew~L. Ginsberg.
\newblock Limited discrepancy search.
\newblock In {\em Proceedings of the 14th international joint conference on
  Artificial intelligence}, volume~1, pages 607--613, 1995.

\bibitem{he2016deep}
Kaiming He, Xiangyu Zhang, Shaoqing Ren, and Jian Sun.
\newblock Deep residual learning for image recognition.
\newblock In {\em Proceedings of the IEEE conference on computer vision and
  pattern recognition}, pages 770--778, 2016.

\bibitem{hoos2011automated}
Holger~H. Hoos.
\newblock Automated algorithm configuration and parameter tuning.
\newblock In {\em Autonomous search}, pages 37--71. Springer, 2011.

\bibitem{hussein2017imitation}
Ahmed Hussein, Mohamed~Medhat Gaber, Eyad Elyan, and Chrisina Jayne.
\newblock Imitation learning: A survey of learning methods.
\newblock {\em ACM Computing Surveys (CSUR)}, 50(2):1--35, 2017.

\bibitem{khalil_learning_2016}
Elias Khalil, Pierre Le~Bodic, Le~Song, George Nemhauser, and Bistra Dilkina.
\newblock Learning to branch in mixed integer programming.
\newblock In {\em Proceedings of the AAAI Conference on Artificial
  Intelligence}, volume~30, 2016.

\bibitem{kingma2014adam}
Diederik~P. Kingma and Jimmy Ba.
\newblock Adam: A method for stochastic optimization.
\newblock {\em arXiv preprint arXiv:1412.6980}, 2014.

\bibitem{kool2018attention}
Wouter Kool, Herke van Hoof, and Max Welling.
\newblock Attention, learn to solve routing problems!
\newblock In {\em International Conference on Learning Representations}, 2019.
\newblock URL: \url{https://openreview.net/forum?id=ByxBFsRqYm}.

\bibitem{kruber2017learning}
Markus Kruber, Marco~E L{\"u}bbecke, and Axel Parmentier.
\newblock Learning when to use a decomposition.
\newblock In {\em International conference on AI and OR techniques in
  constraint programming for combinatorial optimization problems}, pages
  202--210. Springer, 2017.

\bibitem{lecun2015deep}
Yann LeCun, Yoshua Bengio, and Geoffrey Hinton.
\newblock Deep learning.
\newblock {\em Nature}, 521(7553):436--444, 2015.

\bibitem{maas2013rectifier}
Andrew~L. Maas, Awni~Y. Hannun, Andrew~Y. Ng, et~al.
\newblock Rectifier nonlinearities improve neural network acoustic models.
\newblock In {\em Proc. icml}, volume~30, page~3. Atlanta, Georgia, USA, 2013.

\bibitem{mazyavkina2021reinforcement}
Nina Mazyavkina, Sergey Sviridov, Sergei Ivanov, and Evgeny Burnaev.
\newblock Reinforcement learning for combinatorial optimization: A survey.
\newblock {\em Computers \& Operations Research}, 134:105400, 2021.

\bibitem{michel2021minicp}
Laurent Michel, Pierre Schaus, and Pascal van Hentenryck.
\newblock Mini{CP}: a lightweight solver for constraint programming.
\newblock {\em Mathematical Programming Computation}, 13:133--184, 2021.

\bibitem{michel2012activity}
Laurent Michel and Pascal van Hentenryck.
\newblock Activity-based search for black-box constraint programming solvers.
\newblock In {\em International Conference on Integration of Artificial
  Intelligence and Operations Research Techniques in Constraint Programming},
  pages 228--243. Springer, 2012.

\bibitem{4066245}
Marvin Minsky.
\newblock Steps toward artificial intelligence.
\newblock {\em Proceedings of the IRE}, 49(1):8--30, 1961.
\newblock \href {https://doi.org/10.1109/JRPROC.1961.287775}
  {\path{doi:10.1109/JRPROC.1961.287775}}.

\bibitem{DQN}
Volodymyr Mnih, Koray Kavukcuoglu, David Silver, Alex Graves, Ioannis
  Antonoglou, Daan Wierstra, and Martin Riedmiller.
\newblock Playing atari with deep reinforcement learning.
\newblock {\em arXiv preprint arXiv:1312.5602}, 2013.

\bibitem{morabit2021machine}
Mouad Morabit, Guy Desaulniers, and Andrea Lodi.
\newblock Machine-learning--based column selection for column generation.
\newblock {\em Transportation Science}, 55(4):815--831, 2021.

\bibitem{nethercote2007minizinc}
Nicholas Nethercote, Peter~J. Stuckey, Ralph Becket, Sebastian Brand,
  Gregory~J. Duck, and Guido Tack.
\newblock Minizinc: Towards a standard {CP} modelling language.
\newblock In {\em International Conference on Principles and Practice of
  Constraint Programming}, pages 529--543. Springer, 2007.

\bibitem{potvin1996hybrid}
Jean-Yves Potvin, Danny Dub{\'e}, and Christian Robillard.
\newblock A hybrid approach to vehicle routing using neural networks and
  genetic algorithms.
\newblock {\em Applied Intelligence}, 6(3):241--252, 1996.

\bibitem{refalo2004impact}
Philippe Refalo.
\newblock Impact-based search strategies for constraint programming.
\newblock In {\em International Conference on Principles and Practice of
  Constraint Programming}, pages 557--571. Springer, 2004.

\bibitem{treeMDP}
Lara Scavuzzo, Feng~Yang Chen, Didier Ch{\'e}telat, Maxime Gasse, Andrea Lodi,
  Neil Yorke-Smith, and Karen Aardal.
\newblock Learning to branch with tree {MDPs}.
\newblock {\em arXiv preprint arXiv:2205.11107}, 2022.

\bibitem{schaul2015prioritized}
Tom Schaul, John Quan, Ioannis Antonoglou, and David Silver.
\newblock Prioritized experience replay.
\newblock {\em arXiv preprint arXiv:1511.05952}, 2015.

\bibitem{schulman2015trust}
John Schulman, Sergey Levine, Pieter Abbeel, Michael Jordan, and Philipp
  Moritz.
\newblock Trust region policy optimization.
\newblock In {\em International conference on machine learning}, pages
  1889--1897, 2015.

\bibitem{selsam2019guiding}
Daniel Selsam and Nikolaj Bj{\o}rner.
\newblock Guiding high-performance {SAT} solvers with unsat-core predictions.
\newblock In {\em International Conference on Theory and Applications of
  Satisfiability Testing}, pages 336--353. Springer, 2019.

\bibitem{song_learning_2022}
Wen Song, Zhiguang Cao, Jie Zhang, Chi Xu, and Andrew Lim.
\newblock Learning variable ordering heuristics for solving constraint
  satisfaction problems.
\newblock {\em Engineering Applications of Artificial Intelligence},
  109:104603, 2022.

\bibitem{sun2020improving}
Haoran Sun, Wenbo Chen, Hui Li, and Le~Song.
\newblock Improving learning to branch via reinforcement learning.
\newblock In {\em Learning Meets Combinatorial Algorithms at NeurIPS2020},
  2020.

\bibitem{sunehag2017valuedecomposition}
Peter Sunehag, Guy Lever, Audrunas Gruslys, Wojciech~Marian Czarnecki, Vinicius
  Zambaldi, Max Jaderberg, Marc Lanctot, Nicolas Sonnerat, Joel~Z. Leibo, Karl
  Tuyls, and Thore Graepel.
\newblock Value-decomposition networks for cooperative multi-agent learning,
  2017.
\newblock \href {http://arxiv.org/abs/1706.05296} {\path{arXiv:1706.05296}}.

\bibitem{sutton2018reinforcement}
Richard~S. Sutton and Andrew~G. Barto.
\newblock {\em Reinforcement learning: An introduction}.
\newblock MIT press, 2018.

\bibitem{tardivo2022constraints}
Fabio Tardivo and Agostino Dovier.
\newblock Constraints propagation on {GPU}: A case study for alldifferent.
\newblock In {\em Proceedings of the 37th Italian Conference on Computational
  Logic}, 2022.

\bibitem{trott2019keeping}
Alexander Trott, Stephan Zheng, Caiming Xiong, and Richard Socher.
\newblock Keeping your distance: Solving sparse reward tasks using
  self-balancing shaped rewards.
\newblock {\em Advances in Neural Information Processing Systems}, 32, 2019.

\bibitem{van2021learning}
Ronald van Driel, Emir Demirovi{\'c}, and Neil Yorke-Smith.
\newblock Learning variable activity initialisation for lazy clause generation
  solvers.
\newblock In {\em Integration of Constraint Programming, Artificial
  Intelligence, and Operations Research: 18th International Conference, CPAIOR
  2021, Vienna, Austria, July 5--8, 2021, Proceedings 18}, pages 62--71.
  Springer, 2021.

\bibitem{van2016deep}
Hado van Hasselt, Arthur Guez, and David Silver.
\newblock Deep reinforcement learning with double {Q}-learning.
\newblock In {\em Proceedings of the AAAI conference on artificial
  intelligence}, volume~30, 2016.

\bibitem{xia2018learning}
Wei Xia and Roland Yap.
\newblock Learning robust search strategies using a bandit-based approach.
\newblock In {\em Proceedings of the AAAI Conference on Artificial
  Intelligence}, volume~32, 2018.

\bibitem{yu2022combinatorial}
Xin Yu, Thiago Serra, Srikumar Ramalingam, and Shandian Zhe.
\newblock The combinatorial brain surgeon: Pruning weights that cancel one
  another in neural networks.
\newblock In {\em International Conference on Machine Learning}, pages
  25668--25683. PMLR, 2022.

\end{thebibliography}
\end{document}